\documentclass[10pt,journal,letterpaper,compsoc,twocolumn]{IEEEtran}

\usepackage{cite}
\usepackage{graphicx}
\usepackage{psfrag}
\usepackage{subfigure}
\usepackage{url}
\usepackage{stfloats}
\usepackage{amssymb,amsmath}
\usepackage{amsfonts}
\usepackage{array}
\usepackage[ruled]{algorithm2e}

\begin{document}

\title{Fast Online EM for Big Topic Modeling}

\author{Jia~Zeng,~\IEEEmembership{Senior Member,~IEEE}
Zhi-Qiang~Liu
and
Xiao-Qin Cao
\IEEEcompsocitemizethanks{\IEEEcompsocthanksitem
J.~Zeng is with the School of Computer Science and Technology,
Soochow University, Suzhou 215006, China, and the Huawei Noah's Ark Lab, Hong Kong.
E-mail: j.zeng@ieee.org.
\IEEEcompsocthanksitem
Z.-Q.~Liu and X.-Q.~Cao are with the School of Creative Media,
City University of Hong Kong, Hong Kong, China.
}
}

\IEEEcompsoctitleabstractindextext{

\begin{abstract}
The expectation-maximization (EM) algorithm can compute the maximum-likelihood (ML) or maximum a posterior (MAP)
point estimate of the mixture models or latent variable models such as latent Dirichlet allocation (LDA),
which has been one of the most popular probabilistic topic modeling methods in the past decade.
However,
batch EM has high time and space complexities to learn big LDA models from big data streams.
In this paper,
we present a fast online EM (FOEM) algorithm that infers the topic distribution
from the previously unseen documents incrementally with constant memory requirements.
Within the stochastic approximation framework,
we show that FOEM can converge to the local stationary point of the LDA's likelihood function.
By dynamic scheduling for the fast speed and parameter streaming for the low memory usage,
FOEM is more efficient for some lifelong topic modeling tasks than the state-of-the-art online LDA algorithms to
handle both big data and big models (aka, big topic modeling) on just a PC.
\end{abstract}


}

\maketitle

\IEEEpeerreviewmaketitle

\section{Introduction} \label{s1}

Probabilistic topic modeling~\cite{Blei:12} automatically finds word clusters or distributions called topics from a large corpus.
Latent Dirichlet allocation (LDA)~\cite{Blei:03} is one of the most popular topic modeling paradigms,
which has found many important applications in machine learning, computer vision and natural language processing.
From the thematic labeling point of view,
LDA assigns the hidden topic labels (variables) to explain the observed words in document-word matrix~\cite{Zeng:11}.
This labeling process defines a joint probability distribution over the hidden labels and the observed words.
Employing the Bayes' rule,
we can infer the topic labels from the observed words by computing
the posterior distribution of the hidden variables given the observed variables from their joint probability.
Typical batch LDA algorithms include
expectation-maximization (EM)~\cite{Freitas:01},
variational Bayes (VB)~\cite{Blei:03},
Gibbs sampling (GS)~\cite{Griffiths:04},
collapsed variational Bayes (CVB)~\cite{Teh:07,Asuncion:09},
and belief propagation (BP)~\cite{Zeng:11,Zeng:12,Zeng:12a}.

In the big data era,
we need lifelong topic modeling algorithms that can infer a large number of parameters of
LDA from big data streams without ending (aka, big topic modeling).
However,
previous batch algorithms have to sweep repeatedly the entire data set until convergence,
so that they have very high time and space complexities scaling linearly with the number of documents $D$ and the number of topics $K$.
For example,
batch VB~\cite{Blei:03} requires a few days to scan $D = 8,200,000$ PUBMED documents~\cite{Porteous:08}
for $K = 100$ when the number of sweeps $T=100$.
Moreover,
batch VB cannot even fit the entire PUBMED corpus in $4$GB memory of a common PC.
To process big data,
online~\cite{Yao:09,Hoffman:10,Wahabzada:11,Mimno:12,Foulds:13,Zhai:13}
and parallel~\cite{Newman:09,Smola:12,Wang:15,Liu:15} LDA algorithms are two widely used solutions.
Since parallel algorithms depend on expensive parallel hardware,
in this paper,
we focus on online LDA algorithms that require only a constant memory usage
to detect topic distribution shifts as the big data stream flows on just a PC.
Note that though we discuss only data streams in this paper,
online LDA algorithms can process both batch and stream data.
Moreover,
we may parallelize online LDA algorithms in the multi-core or multi-processor environment to simultaneously handle
multiple data streams for a better scalability~\cite{Liu:15}.

Indeed,
big topic modeling has shown potential business values in real-world industrial applications such as search engine,
online advertising systems and churn prediction~\cite{Wang:15,Huang:15}.
As reported by the Linguistic Data Consortium (LDC),
there are millions of vocabulary words in English, Chinese, Spanish, and Arabic.
Taking polysemy and synonyms into consideration,
a rough estimate of the number of word senses is close to the same magnitude of vocabulary words---that is,
around $10^5$ or $10^6$ topics for semantics of small correlated word sets.
Extensive experiments on big search query data confirm that inferring at least $10^5$ topics
can achieve a significant improvement on industrial search engine and online advertising systems~\cite{Wang:15}.
More specifically,
big topic modeling requires to handle the following tasks:
\begin{enumerate}
\item
When the data stream is too large (e.g., $D \ge 10^7$) to fit in memory;
\item
When the number of LDA parameters is too large (e.g., $\ge 10^9$) to fit in memory;
\item
When the number of extracted topics (e.g., $K \ge 10^5$) is very large;
\item
When the vocabulary size (e.g., $W \ge 10^5$) in data streams is very large.
\end{enumerate}
The above four tasks can be categorized broadly into two problems:
{\em big data} and {\em big model}.
The former indicates that the size of data sets is too large to fit in memory,
while the latter means that the number of model parameters is too big to fit in memory.
To handle the big topic modeling tasks,
online algorithms partition a stream of infinite $D \rightarrow \infty$ documents into small minibatches with size $D_s$,
and use the stochastic gradient produced by each minibatch to estimate topic distributions incrementally~\cite{Robbins:51}.
Each minibatch is discarded from the memory after one look.
So,
the memory cost scales linearly with the minibatch size $D_s \ll D$,
where $D_s$ is often a fixed number provided by users.
Recently,
most online LDA algorithms are combinations of the stochastic optimization framework~\cite{Robbins:51}
with batch LDA algorithms like VB, GS, and CVB,
e.g.,
online VB (OVB)~\cite{Hoffman:10},
residual VB (RVB)~\cite{Wahabzada:11},
online GS (OGS)~\cite{Yao:09},
sampled online inference (SOI)~\cite{Mimno:12},
and stochastic CVB (SCVB)~\cite{Foulds:13}.
However,
these algorithms focus mainly on the {\em big data} problem but rarely on the {\em big model} problem.

In this paper,
we propose a novel fast online EM (FOEM) algorithm for big topic modeling tasks on just a single PC.
First,
we derive the EM framework~\cite{Freitas:01,Liu:15} for learning LDA based on~\cite{Dempster:77},
and discuss two online EM (OEM) variants~\cite{Neal:98,Sato:00,Olivier:09,Liang:09} with convergence proofs.
In our conference paper~\cite{Liu:15},
we focus on how to parallelize EM variants for learning LDA in the shared memory environment.
Second,
we speedup OEM called FOEM for big models (i.e., the large number of LDA parameters)
by two novel techniques:
dynamic scheduling~\cite{Zeng:13} and parameter streaming~\cite{Zeng:14}:\footnote{The dynamic scheduling and parameter streaming
techniques appear in our two unpublished/unsubmitted Arxiv papers~\cite{Zeng:13,Zeng:14}.}
\begin{enumerate}
\item
To reduce the time complexity of OEM,
we propose a residual-based dynamic scheduling method,
which selects and updates only responsibilities and parameters of
the subset of topics and documents or vocabulary words
at each iteration to speedup the convergence of OEM.
\item
To reduce the space complexity of OEM,
we propose an I/O-efficient parameter streaming method that loads only a subset of LDA parameters from hard disk into memory for online optimization.
Since all LDA parameters are stored in hard disk or other external storage,
it enjoys a good fault tolerance and can continue to estimate LDA parameters in the lifelong learning environment.
\end{enumerate}
Through these two techniques,
the proposed FOEM can simultaneously solve both {\em big data} and {\em big model }problems within the unified EM framework.
Moreover,
we show that the unified EM framework can explain recent LDA inference algorithms like VB~\cite{Blei:03},
GS~\cite{Griffiths:04}, CVB~\cite{Asuncion:09} and BP~\cite{Zeng:12}.
Experiments on four big data streams confirm that FOEM is significantly faster and more memory-efficient than
the state-of-the-art online LDA algorithms including
OGS~\cite{Yao:09},
OVB~\cite{Hoffman:10},
RVB~\cite{Wahabzada:11},
SOI~\cite{Mimno:12}£¬
and SCVB~\cite{Foulds:13}.
We anticipate that the proposed FOEM can be also extended to compute ML or MAP estimate
of other mixture models and latent variable models~\cite{Murphy:book}.

The rest of this paper is organized as follows.
Section~\ref{s2} derives EM for LDA,
and discusses its relationship to other LDA algorithms.
Section~\ref{s3} presents FOEM by two techniques: dynamic scheduling for fast speed and parameter streaming for low memory usage.
Section~\ref{s4} compares FOEM with several state-of-the-art online LDA algorithms on four real-world text streams.
Finally,
Section~\ref{s5} draws conclusions and envisions future work.

\section{EM Framework for LDA} \label{s2}

\begin{table}[t]
\centering
\caption{Symbols and Notations.}
\begin{tabular}{|l|l|} \hline
$1 \le d \le D$                      &Document index                 \\ \hline
$1 \le w \le W$                      &Word index in vocabulary       \\ \hline
$1 \le k \le K$                      &Topic index                    \\ \hline
$1 \le t \le T$                      &Iteration index                \\ \hline
$1 \le s \le S$                      &Minibatch index                \\ \hline
$NNZ$                                &Number of non-zero elements    \\ \hline
$D_s$                                &Minibatch size             \\ \hline
$\lambda_k K, \lambda_k \in (0,1]$         &The size of topic subsect      \\ \hline
$\lambda_w W, \lambda_w \in (0,1]$         &The size of vocabulary word subset    \\ \hline
$\mu_{w,d}(k)$                       &Responsibility                 \\ \hline
$r_{w,d}(k)$                         &Residual                       \\ \hline
$\mathbf{x}_{W \times D} = \{x_{w,d}\}$     &Document-word matrix           \\ \hline
$\mathbf{z}_{W \times D} = \{z^k_{w,d}\}$   &Topic labels for words         \\ \hline
$\boldsymbol{\theta}_{K \times D}, \hat{\boldsymbol{\theta}}_{K \times D}$          &Document-topic multinomial distribution    \\ \hline
$\boldsymbol{\phi}_{K \times W}, \hat{\boldsymbol{\phi}}_{K \times W}$            &Topic-word multinomial distribution        \\ \hline
$\alpha,\beta$                       &Dirichlet hyperparameters      \\ \hline
$\rho_s$                             &Learning rate                  \\ \hline
\end{tabular}
\label{notation}
\end{table}

\begin{table}[t]
\centering
\caption{Acronyms.}
\begin{tabular}{|l|l||l|l|} \hline
LDA                      &Latent Dirichlet allocation                &BEM &Batch EM\\ \hline
EM                       &Expectation-maximization                   &IEM &Incremental EM\\ \hline
VB                       &Variational Bayes                          &SEM &Stepwise EM\\ \hline
RVB                      &Residual VB                                &OEM  &Online EM \\ \hline
GS                       &Collapsed Gibbs Sampling                   &FOEM &Fast online EM\\ \hline
CVB                      &Collapsed Variational Bayes                &OVB &Online VB\\ \hline
BP                       &Belief Propagation                         &OGS &Online GS\\ \hline
SOI                      &Sampled online inference                   &SCVB &Stochastic CVB\\ \hline
\end{tabular}
\label{acronyms}
\end{table}

\begin{figure*}[t]
\centering
\includegraphics[width=0.8\linewidth]{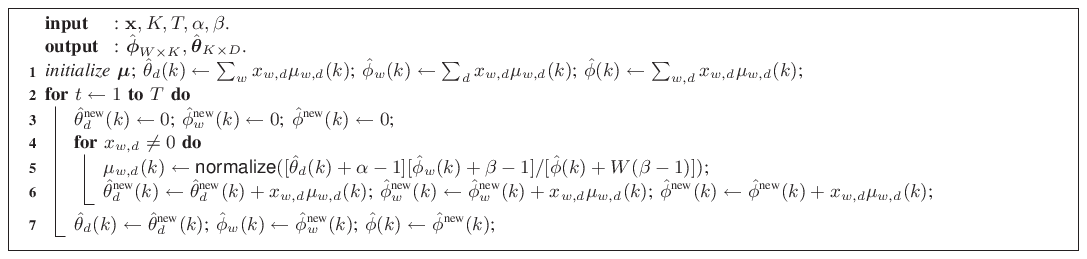}
\caption{Batch EM (BEM) for LDA.}
\label{bem}
\end{figure*}

\begin{figure*}[t]
\centering
\includegraphics[width=0.8\linewidth]{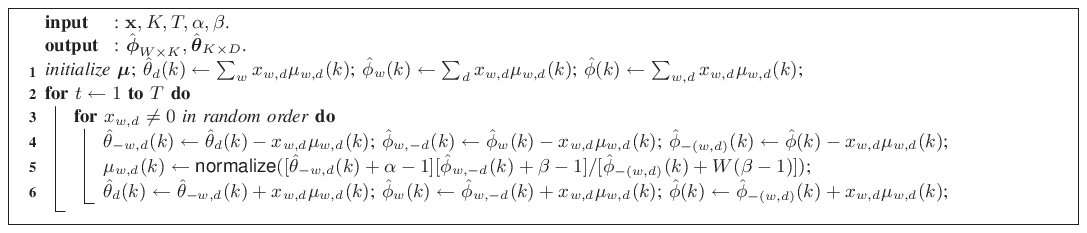}
\caption{Incremental EM (IEM) for LDA.}
\label{iem}
\end{figure*}

\begin{figure*}[t]
\centering
\includegraphics[width=0.8\linewidth]{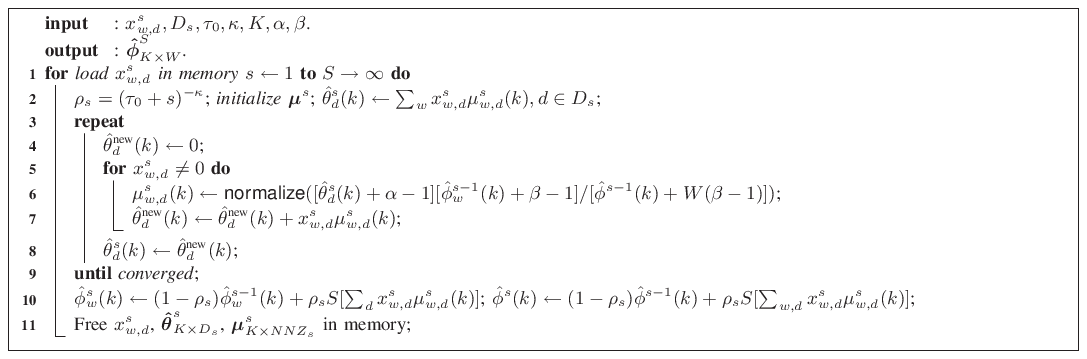}
\caption{Stepwise EM (SEM) for LDA.}
\label{sem}
\end{figure*}

LDA allocates a set of thematic topic labels,
$\mathbf{z} = \{z^k_{w,d}\}$,
to explain non-zero elements in the document-word co-occurrence matrix $\mathbf{x}_{W \times D} = \{x_{w,d}\}$,
where $1 \le w \le W$ denotes the word index in the vocabulary,
$1 \le d \le D$ denotes the document index in the corpus,
and $1 \le k \le K$ denotes the topic index.
Usually,
the number of topics $K$ is provided by users.
The nonzero element $x_{w,d} \ne 0$ denotes the number of word counts at the index $\{w,d\}$.
For each word token $x_{w,d,i} = \{0,1\}, x_{w,d} = \sum_i x_{w,d,i}$,
there is a topic label $z^k_{w,d,i} = \{0,1\}, \sum_{k=1}^K z^k_{w,d,i} = 1, 1 \le i \le x_{w,d}$.
The objective of LDA is to maximize the posterior probability $p(\boldsymbol{\theta},\boldsymbol{\phi}|\mathbf{x},\alpha,\beta)
\propto p(\mathbf{x},\boldsymbol{\theta},\boldsymbol{\phi}|\alpha,\beta)$,
where $\boldsymbol{\theta}_{K \times D}$ and $\boldsymbol{\phi}_{K \times W}$
are two non-negative matrices of multinomial parameters for document-topic and topic-word distributions,
satisfying $\sum_k \theta_d(k) = 1$ and $\sum_w \phi_w(k) = 1$.
Both multinomial matrices are generated by two Dirichlet distributions with hyperparameters $\alpha$ and $\beta$.
For simplicity,
we consider the smoothed LDA with fixed symmetric hyperparameters~\cite{Griffiths:04}.
Although learning asymmetric hyperparameters can enhance the predictive performance~\cite{Wallach:09},
we find that estimating hyperparameters within the EM framework is intractable,
which may be studied in our future work.
Table~\ref{notation} and Table~\ref{acronyms} summarize the important notations and acronyms in this paper.

\subsection{Batch EM (BEM) for LDA} \label{EM}

Batch EM (BEM)~\cite{Dempster:77} maximizes the joint probability of LDA
in terms of multinomial parameter set $\lambda = \{\boldsymbol{\theta},\boldsymbol{\phi}\}$
as follows,
\begin{align} \label{likelihood}
p(\mathbf{x}, \boldsymbol{\theta}, \boldsymbol{\phi}|\alpha, \beta) =
\prod_{w,d,i}\bigg[\sum_k p(x_{w,d,i}=1, z^k_{w,d,i}=1 \notag \\
|\theta_d(k), \phi_w(k))\bigg]\prod_d p(\theta_d(k)|\alpha)\prod_k p(\phi_w(k)|\beta).
\end{align}
Employing the Bayes' rule and the definition of multinomial distributions,
we get the word likelihood,
\begin{align} \label{wordlik}
&p(x_{w,d,i}=1,z^k_{w,d,i}=1|\theta_d(k), \phi_w(k)) = \notag \\
&p(x_{w,d,i}=1|z^k_{w,d,i}=1,\phi_w(k)) \times p(z^k_{w,d,i}=1|\theta_d(k)), \notag \\
&= x_{w,d,i}\phi_w(k)\theta_d(k),
\end{align}
which depends only on the word index $\{w,d\}$ instead of the word token index $i$.
Then,
according to the definition of Dirichlet distributions,
the log-likelihood of~\eqref{likelihood} is
\begin{align} \label{log-likelihood}
&\ell(\lambda) \propto \sum_{w,d,i} x_{w,d,i} \bigg[\log \sum_k \mu_{w,d}(k) \frac{\theta_d(k)\phi_w(k)}{\mu_{w,d}(k)}\bigg] \notag \\
&+ \sum_d\sum_k \log [\theta_d(k)]^{\alpha-1} + \sum_k \sum_w \log[\phi_w(k)]^{\beta - 1},
\end{align}
where $\mu_{w,d}(k)$ is some topic distribution over the word index $\{w,d\}$ satisfying
$\sum_k \mu_{w,d}(k) = 1, \mu_{w,d}(k) \ge 0$.
We observe that $\sum_{w,d,i} [x_{w,d,i} = 1] = \sum_{w,d} x_{w,d}$,
so that we can cancel the word token index $i$ in~\eqref{log-likelihood}.
Because the logarithm is concave,
by Jensen's inequality,
we have
\begin{align} \label{lowerbound}
&\ell(\lambda) \ge \ell(\boldsymbol{\mu}, \lambda) =
\sum_{w,d} \sum_k x_{w,d}\mu_{w,d}(k) \bigg[\log \frac{\theta_d(k)\phi_w(k)}{\mu_{w,d}(k)}\bigg] \notag \\
&+ \sum_d\sum_k \log [\theta_d(k)]^{\alpha-1} + \sum_k \sum_w \log[\phi_w(k)]^{\beta - 1},
\end{align}
which gives the lower bound of log-likelihood~\eqref{log-likelihood}.
The equality holds true if and only if
\begin{align} \label{estep}
\mu_{w,d}(k) \propto \theta_d(k)\phi_w(k).
\end{align}
In EM,
the $K$-length vector $\mu_{w,d}(k)$ is the {\bf responsibility}
that the topic $k$ takes for word index $\{w,d\}$~\cite{Murphy:book}.
For this choice of $\mu_{w,d}(k)$,
Eq.~\eqref{lowerbound} gives a tight lower bound on the log-likelihood~\eqref{log-likelihood} we are trying to maximize.
This is called the E-step in EM~\cite{Dempster:77,Murphy:book}.

In the successive M-step,
we then maximize~\eqref{lowerbound} with respect to parameters to obtain a new setting of $\lambda$.
Since the hyperparameters $\{\alpha,\beta\}$ are fixed,
without loss of generality,
we derive only the M-step update for the parameter $\theta_d(k)$.
There is an additional constraint that $\sum_k \theta_d(k) = 1$
because $\theta_d(k)$ is parameter of a multinomial distribution.
To deal with this constraint,
we construct the Lagrangian from~\eqref{lowerbound} by grouping together only the terms that depend on $\theta_d(k)$,
\begin{align} \label{mstep}
\ell(\theta) = \sum_d\sum_k \bigg[\sum_w x_{w,d}\mu_{w,d}(k) + \alpha - 1 \bigg] \log\theta_d(k) \notag \\
+ \delta(\sum_k \theta_d(k) - 1),
\end{align}
where $\delta$ is the Lagrange multiplier.
Taking derivatives,
we find
\begin{align}
\frac{\partial}{\partial_{\theta_d(k)}}\ell(\theta) = \frac{\sum_w x_{w,d}\mu_{w,d}(k) + \alpha - 1}{\theta_d(k)} + \delta.
\end{align}
Setting this to zero and solving,
we get
\begin{align}
\theta_d(k) = \frac{\sum_w x_{w,d}\mu_{w,d}(k) + \alpha - 1}{-\delta}.
\end{align}
Using the constraint that $\sum_k \theta_d(k) = 1$,
we easily find that $-\delta = \sum_k [\sum_w x_{w,d}\mu_{w,d}(k) + \alpha - 1]$.
We therefore have our M-step update for the parameter $\theta_d(k)$ as
\begin{align} \label{mtheta}
\theta_d(k) = \frac{\hat{\theta}_d(k) + \alpha - 1}{\sum_k \hat{\theta}_d(k) + K(\alpha - 1)}.
\end{align}
where $\hat{\theta}_d(k) = \sum_w x_{w,d}\mu_{w,d}(k)$ is the expected sufficient statistics.
Similarly,
the other multinomial parameter can be estimated by
\begin{align} \label{mphi}
\phi_w(k) = \frac{\hat{\phi}_w(k) + \beta - 1}{\hat{\phi}(k) + W(\beta - 1)},
\end{align}
where $\hat{\phi}_w(k) = \sum_d x_{w,d}\mu_{w,d}(k)$ is the expected sufficient statistics
and we use the notation $\hat{\phi}(k) = \sum_w \hat{\phi}_w(k)$.
Note that the denominator of~\eqref{mtheta} is a constant.
Replacing~\eqref{mtheta} and~\eqref{mphi} into~\eqref{estep},
we obtain the E-step in terms of sufficient statistics,
\begin{align} \label{estep2}
\mu_{w,d}(k) \propto \frac{[\hat{\theta}_d(k) + \alpha - 1]
\times [\hat{\phi}_{w}(k) + \beta - 1]}{\hat{\phi}(k) + W(\beta - 1)},
\end{align}
where EM iterates the E-step and the M-step to refine sufficient statistics $\hat{\theta}_d(k)$ and $\hat{\phi}_{w}(k)$,
which can be normalized to get the multinomial parameters according to~\eqref{mtheta} and~\eqref{mphi}.
Suppose $\lambda^{t-1}$ and $\lambda^t$ are the parameters from two successive iterations $t-1$ and $t$ of EM.
It is easy to prove that
\begin{align} \label{converge}
\ell(\lambda^t) \ge \ell(\boldsymbol{\mu}^{t-1},\lambda^t) \ge \ell(\boldsymbol{\mu}^{t-1}, \lambda^{t-1}) = \ell(\lambda^{t-1}),
\end{align}
which shows that EM always monotonically improves the LDA's log-likelihood~\eqref{log-likelihood} for convergence.
The EM can be also viewed as a coordinate ascent on the lower bound $\ell(\boldsymbol{\mu},\lambda)$,
in which the E-step maximizes it with respect to $\boldsymbol{\mu}$,
and the M-step maximizes it with respect to the multinomial parameter set $\lambda$.
Fig~\ref{bem} summarizes the BEM algorithm for LDA,
where lines $5$ and $6$ are the E-step and M-step, respectively.
BEM sweeps the number of non-zero elements ($NNZ$) in $\mathbf{x}_{W \times D}$ by $K$ times for $1 \le t \le T$ iterations until converged.

\subsection{Online EM (OEM) for LDA}

There are two main online EM (OEM) algorithms~\cite{Murphy:book,Liang:09}:
incremental EM (IEM)~\cite{Neal:98} and stepwise EM (SEM)~\cite{Sato:00,Olivier:09}.
In BEM (Fig.~\ref{bem}),
the M-step is performed until the E-step updates all responsibilities $\mu_{w,d}(k)$,
which slows down the convergence since the updated responsibility of each word in the E-step
does not immediately influence the parameter estimation in the M-step.
This problem motivates IEM.
When compared with BEM~\eqref{estep2},
IEM alternates a single E-step and M-step for each nonzero element $x_{w,d}$ sequentially.
Thus,
the E-step of IEM becomes
\begin{align} \label{estep3}
\mu_{w,d}(k) \propto \frac{[\hat{\theta}_{-w,d}(k) + \alpha - 1]
\times [\hat{\phi}_{w,-d}(k) + \beta - 1]}{\hat{\phi}_{-(w,d)}(k) + W(\beta - 1)}.
\end{align}
The {\em expected sufficient statistics} are
\begin{gather}
\label{stheta}
\hat{\theta}_{-w,d}(k) = \sum_{-w} x_{w,d}\mu_{w,d}(k), \\
\label{sphi}
\hat{\phi}_{w,-d}(k) = \sum_{-d} x_{w,d}\mu_{w,d}(k), \\
\label{sphitot}
\hat{\phi}_{-(w,d)}(k) = \sum_{-(w,d)} x_{w,d}\mu_{w,d}(k),
\end{gather}
where $-w$, $-d$ and $-(w,d)$ denote all word indices except $w$, all document indices except $d$,
and all word indices except $\{w,d\}$.
After the E-step for each word,
the M-step will update the sufficient statistics immediately by adding the updated responsibility $\mu_{w,d}(k)$~\eqref{estep3}
into~\eqref{stheta},~\eqref{sphi} and~\eqref{sphitot}.

Comparing the E-step between BEM and IEM,
we find that the major difference between~\eqref{estep2} and~\eqref{estep3} is that
IEM excludes the current posterior $x_{w,d}\mu_{w,d}(k)$ from sufficient statistics in~\eqref{stheta},~\eqref{sphi} and~\eqref{sphitot}.
Note that CVB0~\cite{Asuncion:09} and asynchronous BP~\cite{Zeng:11,Zeng:13} are equivalent to IEM,
which are also memory-consuming for big data on a single PC.
As far as convergence is concerned,
it is easy to see that IEM can also converge to the local stationary point of LDA's log-likelihood because
\begin{align} \label{converge2}
&\ell(\lambda^t) = \ell(\boldsymbol{\mu}^{t},\lambda^t) \ge \ell(\mu_{w,d}^t,\boldsymbol{\mu}_{-(w,d)}^{t-1},\lambda^t) \notag \\
&\ge \ell(\mu_{w,d}^t,\boldsymbol{\mu}_{-(w,d)}^{t-1}, \lambda^{t-1}) \ge \ell(\boldsymbol{\mu}^{t-1},\lambda^{t-1}) = \ell(\lambda^{t-1}).
\end{align}

Unlike IEM,
SEM takes as input a stream of document-major minibatches,
$x^s_{w,d}, d \in [1, D_s], w \in [1, \infty), s \in [1, S \rightarrow \infty]$,
where $s$ is the index of minibatch and $D_s$ the number of documents in the minibatch.
Note that the minibatch index $s$ and vocabulary word index $w$ can reach infinity
accounting for infinite documents and vocabulary words in the data steam.
Each minibatch of data and local parameters will be freed from memory after one look.
The global topic-word parameter matrix $\boldsymbol{\hat{\phi}}_{K \times W}$ depends on all minibatches,
and thus it is stored entirely in memory by previous online LDA
algorithms~\cite{Yao:09,Hoffman:10,Wahabzada:11,Mimno:12,Zhai:13}.
However,
when $W$ and $K$ are very large,
this matrix is hard to fit in memory referred as the {\em big model} problem.

Theoretically,
SEM~\cite{Sato:00,Olivier:09} combines BEM with the stochastic approximation method,
which achieves convergence to the stationary points of the likelihood function by
interpolating between sufficient statistics based on a learning rate $\rho_s$ satisfying
\begin{align} \label{learningrate}
\rho_s = (\tau_0 + s)^{-\kappa},
\end{align}
where $\tau_0$ is a pre-defined number of mini-batches,
$s$ is the minibatch index and $\kappa \in (0.5, 1]$ is provided by users.
SEM can converge to the local stationary point of LDA's likelihood function
from the online coordinate ascent perspective.
Similar to~\eqref{converge2},
it is easy to observe that
\begin{align} \label{converge3}
\ell(\hat{\phi}^s) = \ell(\boldsymbol{\mu}^{s+1:\infty},\mu^s,\hat{\phi}^s)
\ge \ell(\boldsymbol{\mu}^{s+1:\infty},\mu^{s},\hat{\phi}^{s-1}) \notag \\
\ge \ell(\boldsymbol{\mu}^{s:\infty},\mu^{s-1},\hat{\phi}^{s-1}) = \ell(\hat{\phi}^{s-1}),
\end{align}
where $\boldsymbol{\mu}^{s+1:\infty}$ denotes responsibilities of unseen mini-batches from $s+1$ to $\infty$.
Note that the lower bound~\eqref{converge3} will not touch the log-likelihood~\eqref{log-likelihood}
until all responsibilities for data streams have been updated in~\eqref{lowerbound}.
The inequality~\eqref{converge3} confirms that SEM can improve $\hat{\phi}^s$ to maximize the LDA's log-likelihood~\eqref{log-likelihood}.
Fig.~\ref{sem} shows the SEM algorithm for LDA.
It reads each minibatch $x^s_{w,d}$ into memory and runs BEM (Fig.~\ref{bem}) in lines $4-8$ until $\boldsymbol{\mu}^s$ converged.
Then,
the sufficient statistics $\phi_w^s(k)$ is updated by a linear combination between previous $\phi_w^{s-1}(k)$
and the updated sufficient statistics $\sum_d x^s_{w,d}\mu^s_{w,d}(k)$,
\begin{align} \label{stepweight}
\hat{\phi}_w^s(k) = (1 - \rho_s)\hat{\phi}_w^{s-1}(k) + \rho_s S\bigg[\sum_d x^s_{w,d}\mu^s_{w,d}(k)\bigg],
\end{align}
where $S = D/D_s$ is the scaling coefficient~\cite{Hoffman:10,Mimno:12}.
Finally,
SEM frees $x^s_{w,d}$ and the local parameters $\boldsymbol{\mu}^s$, $\hat{\boldsymbol{\theta}}_{K \times D_s}$ from memory.
Since SEM stores only the subset of data and parameters in memory,
it is easy to process big data stream with a low space complexity.

\subsection{Time and Space Complexities}

\begin{table*}[t]
\centering
\caption{Time and Space Complexities of LDA Inference Algorithms.}
\begin{tabular}{|c|c|c|c|} \hline
&Posterior & Time & Space (Memory)
\\ \hline
BEM (BP)
&$p(\boldsymbol{\theta},\boldsymbol{\phi}|\mathbf{x},\alpha,\beta)$
&$2 \times K \times NNZ$
&$D + 2 \times NNZ + 2 \times K \times (D + W)$
\\ \hline
IEM (CVB0 or BP)
&$p(\boldsymbol{\theta},\boldsymbol{\phi}|\mathbf{x},\alpha,\beta)$
&$2 \times K \times NNZ$
&$D + 2 \times NNZ + K \times (D + NNZ + W)$
\\ \hline
SEM (SCVB)
&$p(\boldsymbol{\theta},\boldsymbol{\phi}|\mathbf{x},\alpha,\beta)$
&$2 \times K \times NNZ$
&$D_s + 2 \times NNZ_s + K \times (D_s + NNZ_s + W)$
\\ \hline
FOEM
&$p(\boldsymbol{\theta},\boldsymbol{\phi}|\mathbf{x},\alpha,\beta)$
&$20 \times NNZ + W_s \times K \log K$
&$D_s + 2 \times NNZ_s + K \times (D_s + NNZ_s + W^*)$
\\ \hline
VB
&$p(\boldsymbol{\theta},\mathbf{z}|\mathbf{x},\boldsymbol{\phi},\alpha,\beta)$
&$2 \times K \times NNZ \times digamma$
&$D + 2 \times NNZ + 2 \times K \times (D + W)$
\\ \hline
GS
&$p(\mathbf{z}|\mathbf{x},\alpha,\beta)$
&$\delta_1 \times K \times ntokens$
&$\delta_2 \times K \times W + 2 \times ntokens$ \\ \hline
CVB
&$p(\boldsymbol{\theta},\boldsymbol{\phi},\mathbf{z}|\mathbf{x},\alpha,\beta)$
&$\delta_3 \times 2 \times K \times NNZ$
&$D + 2 \times NNZ + K \times (2 \times (W + D) + NNZ)$ \\ \hline
\end{tabular}
\label{complexity}
\end{table*}

The time and space complexities of BEM, IEM and SEM are shown in Table~\ref{complexity},
where $K$ is the number of topics,
$D$ the number of documents,
$W$ the vocabulary size,
and $NNZ$ the number of nonzero elements in sparse matrix $\mathbf{x}_{W \times D}$.
Loading document-word sparse matrix $\mathbf{x}_{W \times D}$ in memory
requires around $\mathcal{O}(D + 2 \times NNZ)$ (compressed document-major format) or
$\mathcal{O}(W + 2 \times NNZ)$ (compressed vocabulary-major format) space.

BEM in Fig.~\ref{bem} needs to sweep all non-zero elements several iterations until convergence.
For each element,
it requires $K$ iterations to update and another $K$ iteration to normalize the responsibility~\eqref{estep2}.
So,
BEM's time complexity is around $\mathcal{O}(2 \times K \times NNZ)$.
The time complexities of IEM and SEM are similar to that of BEM because all non-zero elements in document-word matrix
have to be swept $2K$ times.
The only difference is the number of iterations for convergence,
and usually
$T_{SEM} < T_{IEM} < T_{BEM}$.

Besides $\mathbf{x}_{W \times D}$ in memory,
BEM stores four parameter matrices
$\{\boldsymbol{\hat{\theta}}_{K \times D}, \boldsymbol{\hat{\theta}}^\text{new}_{K \times D},
\boldsymbol{\hat{\phi}}_{K \times W}, \boldsymbol{\hat{\phi}}^\text{new}_{K \times W}\}$ for $\mathcal{O}(2 \times K \times (D + W))$ space.
Unlike BEM,
IEM in Fig.~\ref{iem} needs to store the large responsibility matrix $\boldsymbol{\mu}_{K \times NNZ}$.
Storing three full matrices of sufficient statistics (parameters) $\boldsymbol{\mu}_{K \times NNZ}$,
$\boldsymbol{\hat{\theta}}_{K \times D}$ and $\boldsymbol{\hat{\phi}}_{K \times W}$ in~\eqref{mtheta} and~\eqref{mphi}
requires a total of $\mathcal{O}(K \times (D + NNZ + W))$ memory space.
For example,
if $K = 100$,
the responsibility matrix will occupy around $360$GB (using double-precision floating-point format) memory
on the PUBMED data set~\cite{Porteous:08} having $483,450,157$ nonzero elements.
This space is currently too large to be afforded by a single commodity PC.
Finally,
SEM in Fig.~\ref{sem} consumes the least memory usage because it only stores the current minibatch $x^s_{w,d}$
and corresponding local minibatch of parameter matrices
$\{\boldsymbol{\mu}^s_{K \times D_s}, \hat{\boldsymbol{\theta}}^s_{K \times D_s}\}$,
where $D_s \ll D$ and $NNZ_s \ll NNZ$.
Also,
SEM frees the current minibatch and local parameters from memory after one look.
So,
the major memory consumption of SEM is the global topic-word matrix $\hat{\boldsymbol{\phi}}_{K \times W}$,
which scales linearly with the number of topics and the vocabulary size.
This space complexity constrains SEM for some lifelong topic modeling tasks because
the endless data stream often contains the large number of topics and infinite vocabulary words~\cite{Zhai:13}.

\subsection{Performance Measures}

Predictive perplexity is a standard performance measure for different LDA algorithms~\cite{Blei:03,Asuncion:09,Zeng:11},
which evaluates the word log-likelihood for the unseen document-word matrix (similar to the out-matrix prediction in~\cite{Wang:11}).
Let us take BEM in Fig.~\ref{bem} as an example to show how to calculate predictive perplexity.
First,
we randomly partition the data set into training and test sets in terms of documents.
Second,
we estimate $\hat{\phi}$ on the training set by several iterations like $T = 500$ until convergence.
Third,
we randomly partition each document into $80\%$ and $20\%$ subsets on the test set in terms of word tokens.
Fixing $\hat{\phi}$,
we estimate $\hat{\theta}$ on the $80\%$ subset of the test data by $500$ iterations,
and then calculate the predictive perplexity on the rest $20\%$ subset,
\begin{align} \label{pp}
\mathcal{P}=\exp\Bigg\{-\frac{\sum_{w,d}
x_{w,d}^{20\%}\log \big[ \sum_{k}\mu^{20\%}_{w,d}(k) \big]}
{\sum_{w,d} x_{w,d}^{20\%}}\Bigg\},
\end{align}
where the unnormalized $\mu^{20\%}_{w,d}(k)$ on test set is the predicted word likelihood $p(z^{20\%}_{w,d}=k|\boldsymbol{\theta},\boldsymbol{\phi})$,
and the higher likelihood means the better predictive performance.
The multinomial parameters $\{\boldsymbol{\theta},\boldsymbol{\phi}\}$ are the normalized sufficient
statistics $\{\hat{\boldsymbol{\theta}}, \hat{\boldsymbol{\phi}}\}$ in~\eqref{mtheta} and~\eqref{mphi}.
The held-out word tokens or counts in the the $20\%$ subset is denoted by $x_{w,d}^{20\%}$.
The lower predictive perplexity represents a better generalization ability or out-matrix prediction ability~\cite{Wang:11}.
Indeed,
the predictive perplexity~\eqref{pp} is a function of the word log-likelihood~\eqref{wordlik}.
The lower predictive perplexity corresponds to the higher word log-likelihood.
Because EM maximizes the tight lower bound~\eqref{lowerbound} of word log-likelihood,
it minimizes predictive perplexity~\eqref{pp} quickly at each iteration.

\subsection{Relationship to Other LDA Algorithms} \label{s2.4}

VB~\cite{Blei:03} infers the posterior from the full joint probability,
\begin{align}
p(\boldsymbol{\theta},\mathbf{z}|\mathbf{x},\boldsymbol{\phi},\alpha,\beta) =
\frac{p(\mathbf{x},\mathbf{z},\boldsymbol{\theta},\boldsymbol{\phi}|\alpha,\beta)}
{p(\mathbf{x},\boldsymbol{\phi}|\alpha,\beta)}.
\end{align}
This posterior means that if we learn the topic-word distribution $\boldsymbol{\phi}$ from training data,
we want to infer the best $\{\boldsymbol{\theta},\mathbf{z}\}$ from unseen test data given $\boldsymbol{\phi}$,
i.e.,
for the best generalization performance.
However,
computing this posterior is intractable because the denominator contains intractable integration,
$\int_{\boldsymbol{\theta},\mathbf{z}}p(\mathbf{x},\mathbf{z},\boldsymbol{\theta},\boldsymbol{\phi}|\alpha,\beta)$.
So,
VB infers an approximate variational posterior based on the variational EM algorithm~\cite{Murphy:book}:
\begin{itemize}
\item
Variational E-step:
\small
\begin{gather} \label{vestep}
\mu_{w,d}(k) \propto
\frac{\exp[\Psi(\hat{\theta}_d(k) + \alpha)]\exp[\Psi(\hat{\phi}_w(k) + \beta)]}{\exp[\Psi(\sum_w [\hat{\phi}_w(k) + \beta])]}, \\
\hat{\theta}_d(k) = \sum_w x_{w,d}\mu_{w,d}(k).
\end{gather}
\normalsize
\item
Variational M-step:
\begin{align}
\hat{\phi}_w(k) = \sum_d x_{w,d}\mu_{w,d}(k).
\end{align}
\end{itemize}
In variational E-step,
we update $\mu_{w,d}(k)$ and $\hat{\theta}_d(k)$ until convergence,
which makes the variational posterior approximate
the true posterior $p(\boldsymbol{\theta},\mathbf{z}|\mathbf{x},\boldsymbol{\phi},\alpha,\beta)$
by minimizing the Kullback-Leibler (KL) divergence between them.
In the variational M-step,
we update $\hat{\phi}_w(k)$ to maximize the variational posterior.
Here,
we use the notation $\hat{\phi}(k) = \sum_w \hat{\phi}_w(k)$ for the denominator in~\eqref{vestep}.
Normalizing $\{\hat{\boldsymbol{\theta}},\hat{\boldsymbol{\phi}}\}$ yields the multinomial parameters $\{\boldsymbol{\theta},\boldsymbol{\phi}\}$.
However,
the variational posterior cannot touch the true posterior for inaccurate solutions~\cite{Blei:03}.
In addition,
the calculation of exponential digamma function $\exp[\Psi(\cdot)]$ is computationally complicated.
As shown in Table~\ref{complexity},
the time complexity of VB for one iteration is $\mathcal{O}(2 \times K \times NNZ \times digamma)$,
where $digamma$ is the computing time for exponential digamma function,
and $NNZ$ is the number of nonzero elements in document-word sparse matrix.
For each nonzero element,
we need $K$ iterations for variational E-step and $K$ iterations for normalizing $\mu_{w,d}(k)$.
Besides the data size $D + 2 \times NNZ$,
the space complexity of VB is $\mathcal{O}(2 \times K \times (D+W))$ for two multinomial parameters
and temporary storage for variational M-step.
Unlike FOEM using time-efficient IEM inference with stochastic gradients,
OVB~\cite{Hoffman:10} and its extension~\cite{Mandt:14}
combine VB inference with unbiased and biased stochastic gradients,
respectively.

In contrast to VB,
the collapsed GS~\cite{Griffiths:04} algorithm infers the posterior for each word token by integrating out the multinomial parameters
$\{\boldsymbol{\theta},\boldsymbol{\phi}\}$,
\begin{align}
p(\mathbf{z}|\mathbf{x},\alpha,\beta) = \frac{p(\mathbf{x},\mathbf{z}|\alpha,\beta)}{p(\mathbf{x}|\alpha,\beta)}
\propto p(\mathbf{x},\mathbf{z}|\alpha,\beta).
\end{align}
This posterior means that we want to find the best topic labeling configuration $\mathbf{z}$ given the observed words $\mathbf{x}$.
The total number of word tokens,
$ntokens = \sum_{w,d} x_{w,d}$.
Because the multinomial parameters $\{\boldsymbol{\theta},\boldsymbol{\phi}\}$ have been integrated out,
the best labeling configuration $\mathbf{z}$ is insensitive to the variation of $\{\boldsymbol{\theta},\boldsymbol{\phi}\}$.
Maximizing the joint probability $p(\mathbf{x},\mathbf{z}|\alpha,\beta)$ is intractable
(i.e., there are $K^{ntokens}$ configurations that increase exponentially),
an approximate inference called Markov chain Monte Carlo (MCMC) EM~\cite{Murphy:book} is used as follows:
\begin{itemize}
\item
MCMC E-step:
\begin{gather}
\label{mcmcestep}
\mu_{w,d,i}(k) \propto
\frac{[\hat{\theta}^{-z^{k,old}_{w,d,i}}_d(k) + \alpha][\hat{\phi}^{-z^{k,old}_{w,d,i}}_w(k) + \beta]}
{\sum_w [\hat{\phi}^{-z^{k,old}_{w,d,i}}_w(k) + \beta]}, \\
\text{Random Sampling} \; z^{k,new}_{w,d,i} = 1 \; \text{from} \; \mu_{w,d,i}(k).
\end{gather}
\item
MCMC M-step:
\begin{gather}
\hat{\theta}_d(k) = \hat{\theta}^{-z^{k,old}_{w,d,i}}_d(k) + z^{k,new}_{w,d,i}, \\
\hat{\phi}_w(k) = \hat{\phi}^{-z^{k,old}_{w,d,i}}_w(k) + z^{k,new}_{w,d,i}.
\end{gather}
\end{itemize}
In the MCMC E-step,
GS infers the topic posterior per word token,
$\mu_{w,d,i}(k) = p(z^{k,new}_{w,d,i}=1|\mathbf{z}^{k,old}_{w,d,-i},\mathbf{x},\alpha,\beta)$,
and randomly samples a new topic label $z^{k,new}_{w,d,i} = 1$ from this posterior.
The notation $-z^{k,old}_{w,d,i}$ means excluding the old topic label from the corresponding matrices
$\{\hat{\boldsymbol{\theta}}, \hat{\boldsymbol{\phi}}\}$.
In the MCMC M-step,
GS updates immediately $\{\hat{\boldsymbol{\theta}},\hat{\boldsymbol{\phi}}\}$ by the new topic label of each word token.
In this sense,
GS can be viewed as an incremental algorithm that learns parameters by processing data point sequentially.
In Table~\ref{complexity},
the time complexity of GS for one iteration is $\mathcal{O}(\delta_1 \times K \times ntokens)$,
where $\delta_1 \ll 2$.
The reason is that we require $K$ iterations in MCMC E-step and less $K$ iterations for normalizing $\mu_{w,d,i}(k)$.
According to sparseness of $\mu_{w,d,i}(k)$,
efficient sampling techniques~\cite{Porteous:08,Yao:09,Li:14} can make $\delta_1$ even smaller.
Practically,
when $K$ is larger than $1000$,
$\delta_1 \approx 0.05$.
Generally,
we do not need to store $\hat{\boldsymbol{\theta}}_{K \times D}$ in memory because $\mathbf{z}$ can recover $\hat{\boldsymbol{\theta}}_{K \times D}$.
So,
besides the data size $ntokens$,
the space complexity of GS is $\mathcal{O}(\delta_2 \times K \times W + ntokens)$ because $\hat{\boldsymbol{\phi}}_{K \times W}$
can be compressed due to sparseness.
When $K$ is larger than $1000$,
$\delta_2 \approx 0.8$.
Note that all parameters in GS are stored in integer type,
saving around half memory space when compared with the double type used by other LDA inference algorithms.
Similar to SEM,
OGS~\cite{Yao:09} combines sparse GS (SGS) with stochastic gradients for online topic modeling.
SOI~\cite{Mimno:12} is a hybrid of OVB and OGS algorithms for sparseness of responsibilities.

Unlike VB and GS,
CVB~\cite{Teh:07} infers the complete posterior given the observed data $\mathbf{x}$,
\begin{align}
p(\boldsymbol{\theta},\boldsymbol{\phi},\mathbf{z}|\mathbf{x},\alpha,\beta) \propto
p(\mathbf{x},\mathbf{z},\boldsymbol{\theta},\boldsymbol{\phi}|\alpha,\beta).
\end{align}
Maximizing this posterior means that we want to obtain the best combination of multinomial parameters $\{\boldsymbol{\theta}, \boldsymbol{\phi}\}$
for the best topic labeling configuration $\mathbf{z}$.
However,
inference of this posterior is intractable so that the Gaussian approximation is used~\cite{Teh:07}.
In this sense,
CVB optimizes an approximate LDA model,
which may not achieve the best topic modeling accuracy.
The variational E-step and M-step in CVB are similar to those in GS.
The main difference is that the variational E-step
requires multiplying an exponential correction factor containing variance update for each nonzero element rather than each word token.
In Table~\ref{complexity},
the time complexity of CVB is $\mathcal{O}(\delta_3 \times 2 \times K \times NNZ)$,
where $\delta_3 > 1$ denotes the additional cost for calculating exponential correction factor.
Besides the data size $D + 2 \times NNZ$,
the space complexity of CVB is $K \times (2 \times (W + D) + NNZ)$ because it needs to
store one copy of matrix $\mu_{K \times NNZ}$,
and two copies of matrices $\hat{\boldsymbol{\theta}}_{K \times D}$ and $\hat{\boldsymbol{\phi}}_{K \times W}$ in memory
(one for the original and the other for the variance)~\cite{Teh:07,Asuncion:09}.
SCVB~\cite{Foulds:13} slightly changes the zero-order approximation of CVB within the stochastic optimization framework,
which is equivalent to SEM.

We advocate the standard EM~\cite{Freitas:01} algorithm that infers the posterior
by integrating out the topic labeling configuration $\mathbf{z}$,
\begin{align} \label{bp}
p(\boldsymbol{\theta},\boldsymbol{\phi}|\mathbf{x},\alpha,\beta)
= \frac{p(\mathbf{x},\boldsymbol{\theta},\boldsymbol{\phi}|\alpha,\beta)}
{p(\mathbf{x}|\alpha,\beta)} \propto
p(\mathbf{x},\boldsymbol{\theta},\boldsymbol{\phi}|\alpha,\beta).
\end{align}
Unlike the posteriors of VB and GS,
this posterior means that we want to find the best parameters $\{\boldsymbol{\theta},\boldsymbol{\phi}\}$ given observations $\mathbf{x}$,
no matter what topic labeling configuration $\mathbf{z}$ is.
To this end,
we integrate out the labeling configuration $\mathbf{z}$ in full joint probability,
and use the standard EM algorithm~\cite{Dempster:77} to optimize this objective~\eqref{bp}.
In the E-step,
EM infers the responsibility
$\mu_{w,d}(k)$ conditioned on parameters $\{\hat{\boldsymbol{\theta}},\hat{\boldsymbol{\phi}}\}$.
In the M-step,
EM updates parameters $\{\hat{\boldsymbol{\theta}},\hat{\boldsymbol{\phi}}\}$ based on the inferred responsibility $\mu_{w,d}(k)$.
Unlike VB,
EM can touch the true posterior distribution $p(\boldsymbol{\theta},\boldsymbol{\phi}|\mathbf{x},\alpha,\beta)$ in the E-step for maximization.
When $\alpha = \beta = 1$,
the Dirichlet distribution becomes the uniform distribution,
which implies that the Dirichlet prior does not constrain the underlying multinomial distribution so that LDA reduces to PLSA~\cite{Hofmann:01}.
In this situation,
we see that BEM in Fig.~\ref{bem} becomes the standard EM algorithm for PLSA without Dirichlet priors.

In the past decade,
VB, GS and CVB have been three main inference algorithms in LDA literatures,
while EM has been rarely discussed and used in learning LDA.
We show two main reasons to use EM:
\begin{enumerate}
\item
\textbf{EM yields a high topic modeling accuracy} measured by predictive perplexity~\eqref{pp},
which is a function of multinomial parameters $\{\boldsymbol{\theta},\boldsymbol{\phi}\}$.
EM infers the best multinomial parameters by maximizing the posterior probability $p(\boldsymbol{\theta},\boldsymbol{\phi}|\mathbf{x},\alpha,\beta)$.
In contrast,
VB, GS and CVB produce relatively higher predictive perplexity than EM because they infer different posteriors.
We see that the synchronous BP~\cite{Zeng:11} is equivalent to BEM (as shown in Table~\ref{complexity}),
which has been confirmed empirically to produce lower predictive perplexity.
\item
\textbf{EM converges faster.}
Convergence analysis shows that all these EM algorithms can converge to the local maximum of LDA's objective function,
because in the E-step the lower-bound can touch the true posterior,
i.e.,
the equality holds in Jensen's inequality~\eqref{lowerbound}.
We see that the zero-order approximation of CVB called CVB0~\cite{Asuncion:09} and the asynchronous BP~\cite{Zeng:11,Zeng:12}
are equivalent to IEM and SCVB~\cite{Foulds:13} is equivalent to SEM (as shown in Table~\ref{complexity}),
which have been confirmed empirically to converge faster than VB, GS and CVB.
Also,
online belief propagation (OBP) for PLSA~\cite{Ye:13} is a special SEM algorithm when $\alpha=\beta=1$.
\end{enumerate}

\section{Fast OEM (FOEM) for LDA} \label{s3}

\begin{figure*}[t]
\centering
\includegraphics[width=0.8\linewidth]{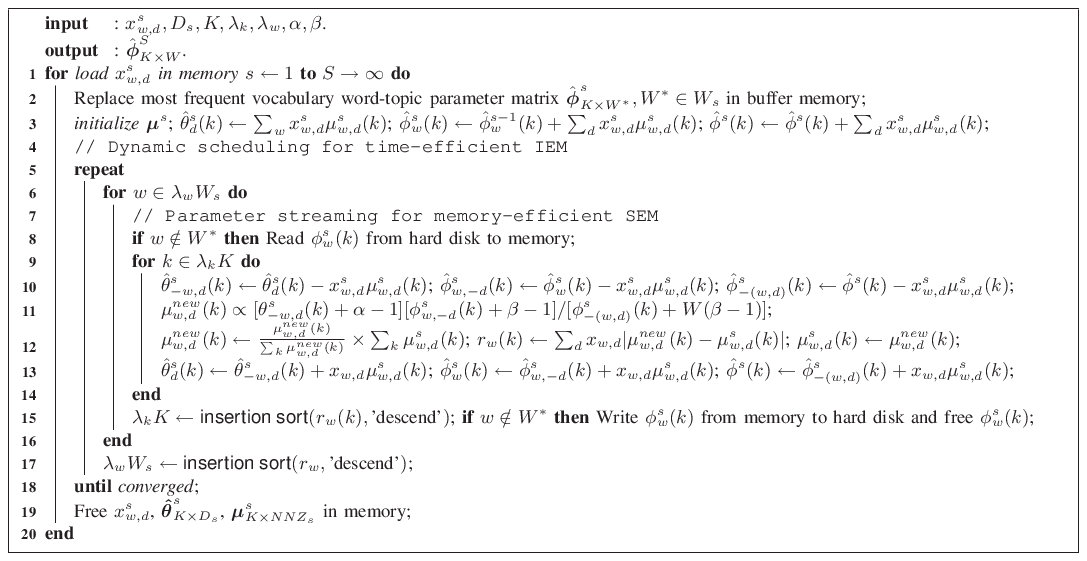}
\caption{Fast online EM (FOEM) for LDA.}
\label{foem}
\end{figure*}

Although SEM is able to process big data streams due to the least memory usage in Table~\ref{complexity},
it still has two scalability issues for big topic modeling tasks discussed in Section~\ref{s1}.
First,
the time complexity scales linearly with the number of topics $K$.
Second,
the space complexity also scales linearly with the number of topics $K$ and the vocabulary size $W$.
For example,
if we extract $K=10^4$ topics from the PUBMED data set ($D=8,200,000$, $W=141,043$, $NNZ=483,450,157$)~\cite{Porteous:08} using SEM,
we require at least $300$ hours to sweep the entire data stream,
and at least $10$GB memory to store the topic-word matrix $\boldsymbol{\hat{\phi}}_{K \times W}$,
which is often unaffordable on just a PC.
To reduce both time and space complexities of SEM,
we propose the fast online EM (FOME) algorithm based on dynamic scheduling and parameter streaming:
\begin{itemize}
\item
We propose a residual-based dynamic scheduling method that can reduce the time complexity of IEM in Table~\ref{complexity} to
$\mathcal{O}(20 \times NNZ + W_s \times K \log K)$.
This time-efficient IEM's time complexity is insensitive to the number of topics $K$.\footnote{The time-efficient
IEM is equivalent to the active belief propagation (ABP) in our unpublished/unsubmitted Arxiv paper~\cite{Zeng:13}.}
\item
We propose a parameter streaming method that can reduce the space complexity (memory usage) of SEM in Table~\ref{complexity}
to $\mathcal{O}(D_s + 2 \times NNZ_s + K \times (D_s + NNZ_s + W^*))$,
where $W^* \ll W$ is a fixed buffer size parameter provided by users.
So,
this memory-efficient SEM's space complexity is also insensitive to the number of topics $K$.
\item
We combine together the time-efficient IEM and the memory-efficient SEM to be the FOEM algorithm for big topic modeling tasks on just a PC.
More specifically,
in Fig.~\ref{sem},
memory-efficient SEM replaces BEM by time-efficient IEM,
which composes the FOEM algorithm.\footnote{FOEM is equivalent to online belief propagation
(OBP) proposed in our unpublished/unsubmitted Arxiv paper~\cite{Zeng:14}.}
\end{itemize}
Fig.~\ref{foem} shows the proposed FOEM algorithm,
which stores in memory only the partial global topic-word matrix $\hat{\boldsymbol{\phi}}^s_{K \times W^*}, W^* \in W_s$,
where $W^*$ is the buffer size and $W_s$ is the vocabulary size of the current minibatch $x^s_{w,d}$ (line $2$).
Then,
FOEM randomly initializes the local responsibilities $\mu^s_{w,d}$ and accumulates them on
corresponding local parameters $\hat{\theta}^s_d(k)$ and $\hat{\phi}^s_w(k)$ (line $3$).
Based on these parameters,
FOEM iteratively performs dynamic scheduling that selects the most important topics for updating at each iteration until converged (lines $5-18$).
If the vocabulary word is not in the buffer,
FOEM uses parameter streaming that reads/writes corresponding parameters from hard disk (lines $8$ and $15$).
More details will be explained in the next subsections.

Unlike SEM in Fig.~\ref{sem},
FOEM does not explicitly use the linear combination of previous global topic-word matrix $\phi^{s-1}$
and updated sufficient statistics $\sum_d x^s_{w,d}\mu^s_{w,d}(k)$ in Eq.~\eqref{stepweight}.
When the parameters $\tau_0 = 0,\kappa = -1$ in Eq.~\eqref{learningrate},
the learning rate $\rho_s = 1/s$ in Eq.~\eqref{stepweight2},
which also satisfies the Robbins-Monro condition~\cite{Robbins:51}.
Through proper normalization~\cite{Liang:09},
Eq.~\eqref{stepweight} can be re-written as
\begin{align} \label{stepweight2}
\hat{\phi}^{s} = (1 - 1/s)\hat{\phi}^{s-1} + \frac{S}{s}\sum_d x^s_{w,d}\mu^s_{w,d}(k),
\end{align}
where the scaling coefficients $(1 - 1/s)$ and $S/s$ on $\hat{\phi}^{s-1}$ and $\sum_d x^s_{w,d}\mu^s_{w,d}(k)$
can be canceled when $s \rightarrow \infty$.
In this way,
we may efficiently calculate the sufficient statistics $\sum_d x^s_{w,d}\mu^s_{w,d}(k)$ of each minibatch,
and accumulate them to topic-word matrix as shown in Fig.~\ref{foem} (line $3$).
As a result,
FOEM can also converge to the local maximum of the LDA's log-likelihood with the learning rate $1/s$
within the stochastic optimization framework~\cite{Robbins:51}.

\subsection{Dynamic Scheduling} \label{s3.1}

\begin{figure}[t]
\centering
\includegraphics[width=0.3\linewidth]{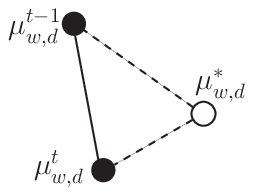}
\caption{Time-efficient IEM: Dynamic scheduling minimizes the largest lower bound first by sorting residuals.}
\label{trianglefig}
\end{figure}

IEM often converges faster than BEM~\cite{Liang:09}.
According to~\eqref{converge2},
the responsibilities,
$\boldsymbol{\mu}^t = \{\mu^t_{1,1}, \dots, \mu^t_{W,D}\}$,
will converge to a set of fixed-points,
$\boldsymbol{\mu}^* = \{\mu^*_{1,1}, \dots, \mu^*_{W,D}\}$.
To speed up convergence,
we use dynamic scheduling that selects to first update the responsibility $\mu_{w,d}$ with the largest distance
$\|\mu^{t}_{w,d} - \mu^*_{w,d}\|$ or $\|\mu^{t-1}_{w,d} - \mu^*_{w,d}\|$,
which will efficiently influence other responsibilities.
However,
we cannot directly measure the distance between a current responsibility and its unknown fixed-point value.
Alternatively,
we can derive a lower bound on this distance that can be calculated easily.
Using the triangle inequality,
we get
\begin{align} \label{triangle}
\|\mu^{t}_{w,d} - \mu^{t-1}_{w,d}\| \le \|\mu^{t}_{w,d} - \mu^*_{w,d}\| + \|\mu^{t-1}_{w,d} - \mu^*_{w,d}\|.
\end{align}
Fig.~\ref{trianglefig} shows the triangle inequality of responsibilities.
In dynamic scheduling,
we minimize the largest lower bound $\|\mu^{t}_{w,d} - \mu^{t-1}_{w,d}\|$ in higher priority,
which defines the responsibility residual between two successive iterations $t$ and $t-1$,
\begin{align} \label{residual}
r^t_{w,d}(k) = x_{w,d}\|\mu^t_{w,d}(k) - \mu^{t-1}_{w,d}(k)\|,
\end{align}
where $x_{w,d}$ is the number of word counts and we choose the $L_1$ norm.
The residual $r^t_{w,d}(k) \rightarrow 0$ as $t \rightarrow \infty$,
which implies the convergence of IEM.

The computational cost of sorting~\eqref{residual} is expensive such as $\mathcal{O}(NNZ \times K\log K)$
because the number of non-zero residuals $r_{w,d}(k)$ is very large in the document-word matrix.
In practice,
we turn to sorting the accumulated residuals at the vocabulary word dimension,
\begin{gather}
\label{residual1}
r_w(k) = \sum_d r_{w,d}(k), \\
\label{residual2}
r_w = \sum_k r_w(k).
\end{gather}
which can be updated during responsibility update at a negligible computational cost.
The time complexity of sorting~\eqref{residual1} in descending order is at most $\mathcal{O}(W_s \times K\log K)$
and sorting~\eqref{residual2} is at most $\mathcal{O}(W_s\log W_s)$.
In each minibatch $x^s_{w,d}$,
the vocabulary size $W_s$ is a constant independent of the number of documents $D_s$.

The time-efficient IEM is a sublinear algorithm of IEM.
Updating and normalizing responsibility~\eqref{estep2} takes $2K$ iterations.
When $K$ is large,
for example, $K \ge 10^4$,
the total number of $2K$ iterations is computationally large to update each responsibility.
Fortunately,
the responsibility vector $\mu_{w,d}(k)$ is very sparse~\cite{Porteous:08,Yao:09,Li:14} when $K$ is large.
From residuals $r_w(k)$ in~\eqref{residual1},
time-efficient IEM selects only a subset of topics with size $\lambda_k K$
having top residuals $r_w(k)$ for responsibility updating and normalization at each learning iteration,
where $\lambda_k \in (0, 1]$ is the ratio parameter provided by the user.
For the selected $\lambda_kK$ topics,
we need to normalize the local responsibilities by
\begin{align} \label{topicmessage}
\hat{\mu}^t_{w,d}(k) = \frac{\mu^t_{w,d}(k)}{\sum_{k}\mu^t_{w,d}(k)} \times \sum_k\hat{\mu}_{w,d}^{t-1}(k), k \in \lambda_k K,
\end{align}
where $\hat{\mu}_{w,d}^{t-1}$ is the normalized responsibility in the previous iteration,
$\hat{\mu}^t_{w,d}(k)$ is the normalized responsibility in the current iteration,
and $\mu^t_{w,d}(k)$ is the unnormalized responsibility updated according to~\eqref{estep3}.
In this way,
we need only $\lambda_k K$ iterations to avoid calculating
the normalization factor $Z = \sum_k \mu_{w,d}(k)$ with $K$ iterations.
Therefore,
time-efficient IEM consumes only $2\lambda_k K$ iterations for responsibility updating and normalization,
where $2 \lambda_k K \ll 2K$.
Furthermore,
when $W_s$ is large,
the time-efficient IEM selects a subset of vocabulary words of size $\lambda_w W_s$,
where $\lambda_w \in (0, 1]$.
Obviously,
the smaller the $\{\lambda_k, \lambda_w\}$ the faster the time-efficient IEM at each iteration.
When $\{\lambda_k = \lambda_w = 1\}$,
the time-efficient IEM becomes the standard IEM in Fig.~\ref{iem}.

Fig.~\ref{foem} summarizes the time-efficient IEM algorithm in FOEM.
In the first iteration (not shown in Fig.~\ref{foem}),
FOEM does not use dynamic scheduling and scans the entire non-zero elements and topics in the minitach $x^s_{w,d}$,
which also initializes and updates the residual matrices $r_w(k)$ and $r_w$.
In the successive iterations (lines $5-18$),
FOEM sorts the residuals and selects the subset of topics $\lambda_k K$ and vocabulary words $\lambda_w W_s$ for updating.
In the meanwhile,
it refines the residuals $r_w(k)$ and $r_w$ at each iteration for dynamic scheduling in the next iteration until converged.
Finally,
at the end of each iteration,
FOEM checks if the training perplexity at successive iterations is less than a predefined threshold
(e.g., $\Delta \mathcal{P} = 10$) to break the loop.
According to~\cite{Zeng:13},
the time-efficient IEM is significantly faster and more accurate than other state-of-the-art batch LDA algorithms.
In this paper,
we set $\{\lambda_k, \lambda_w\}$ parameters of the time-efficient IEM algorithm in FOEM as follows:
\begin{enumerate}
\item
We fix $\lambda_w = 1$ and $\lambda_k K = 10$ because in real-world applications
each vocabulary word is often associated with no more than $10$ topics at each iteration.
In this way,
the runtime of time-efficient IEM is insensitive to the number of topics $K$ except for the sorting time $W_s \times K \log K$.
In this sense,
the time complexity of FOEM becomes $\mathcal{O}(20 \times NNZ + W_s \times K \log K)$.
\item
We adopt the {\em partial sorting} technique for top $\lambda_k K = 10$ largest elements,
which is more efficient than complete sorting and retains almost the same topic modeling accuracy.
In practice,
partial sorting time can be neglected if the responsibility vector is in nearly sorted order.
\end{enumerate}
The time-efficient IEM differs from RVB~\cite{Wahabzada:11} though they both use the residual-based dynamical scheduling techniques.
First,
it uses a more efficient sorting method while RVB uses a relatively complicated sampling technique for dynamical scheduling.
Second,
it can simultaneously schedule vocabulary words and topics for the maximum speedup effects,
while RVB schedules only mini-batches of documents.
Finally,
RVB uses the residuals of document-topic $\hat{\boldsymbol{\theta}}$ parameters instead of responsibilities $\boldsymbol{\mu}$
(the former is the lower bound of the latter),
which may lower the scheduling efficiency.

\subsection{Parameter Streaming} \label{3.2}

\begin{figure}[t]
\centering
\includegraphics[width=1.0\linewidth]{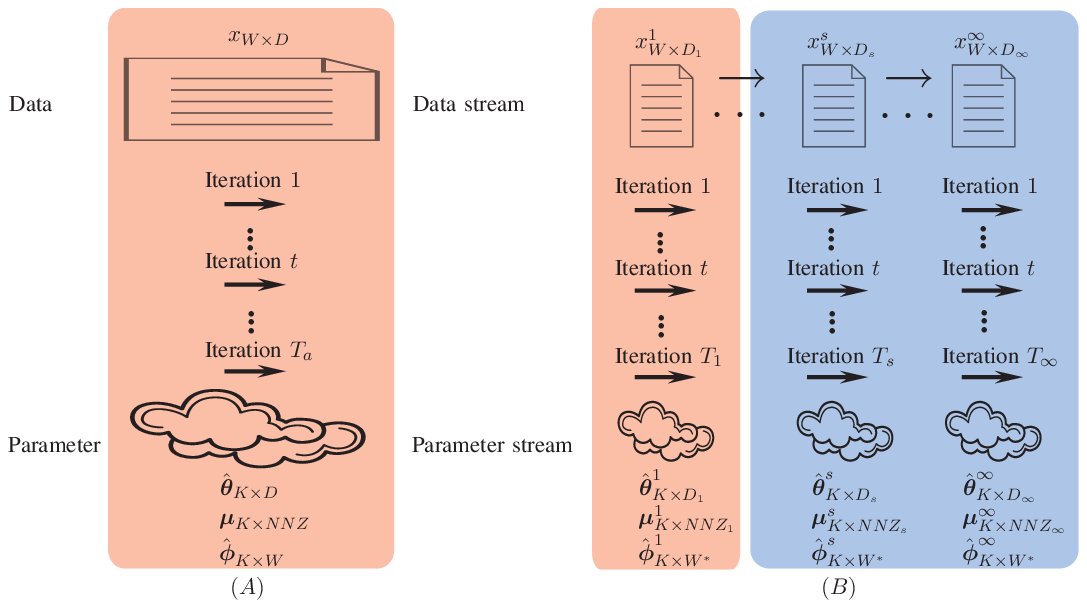}
\caption{(A) IEM and (B) FOEM infer LDA parameters from big data.
Red color plane denotes in-memory computation.
Blue color plane denotes that both data and parameters are stored in secondary storage (disk).}
\label{pstreaming}
\end{figure}

Fig.~\ref{pstreaming} shows how FOEM uses the parameter streaming to reduce
the space complexity of time-efficient IEM in case of both big data and big model.
In Fig.~\ref{pstreaming}A,
time-efficient IEM loads the entire document-word matrix $\mathbf{x}_{W \times D}$ and initializes all LDA parameter matrices such as
$\boldsymbol{\mu}_{K \times NNZ}$,
$\boldsymbol{\hat{\theta}}_{K \times D}$ and $\boldsymbol{\hat{\phi}}_{K \times W}$ in memory denoted by red color plane.
In Fig.~\ref{pstreaming}B,
FOEM loads sequentially in memory only each minibatch of sub-matrix $\mathbf{x}^s_{W_s \times D_s}$
and initializes local parameter matrices $\boldsymbol{\mu}^s_{K \times NNZ_s}$
and $\boldsymbol{\hat{\theta}}^s_{K \times D_s}$ denoted by red color plane,
which will be freed after one look.
If the global matrix $\boldsymbol{\hat{\phi}}_{K \times W_s}$ is very large,
FOEM loads only a subset of needed columns in memory as a parameter stream.
All other minibatches and the global topic-word matrix
$\boldsymbol{\hat{\phi}}_{K \times W}$ are stored in secondary storage (hard disk) denoted by blue color plane.
Since FOEM searches stochastic gradients for each small minibatch,
it consumes less number of iterations until convergence.

Fig.~\ref{foem} summarizes the parameter streaming technique.
To make efficient I/O from disk to memory,
we re-organize each incoming minibatch $x^s_{w,d}$ as a vocabulary-major sparse matrix.
So,
we read and write $w$th column of $\boldsymbol{\hat{\phi}}_{K \times W}$ only once at each iteration of FOEM (lines $8$ and $15$).
We also replace frequently visited columns (vocabulary words) of $\boldsymbol{\hat{\phi}}^s_{K \times W^*}$ in buffer (line $2$) for each minibatch,
which further reduces the read and write frequency of columns in $\boldsymbol{\hat{\phi}}^s_{K \times W_s}$.
When a new vocabulary word is met,
we increment the vocabulary size by one,
$W \leftarrow W + 1$,
in~\eqref{estep3}.
In this way,
FOEM can possibly process both infinite documents and vocabulary words in the data stream without ending.
Incrementing the vocabulary size implies that the topic distribution $\phi$
are generated by a Dirichlet distribution with increasing dimensions.
However,
it does not change the responsibility update~\eqref{estep3} very much when $W$ is large.
As a result,
our heuristic by incrementing vocabulary size works well in the LDA framework.
More complicated methods using Dirichlet processes to handle infinite vocabulary size can be found in~\cite{Zhai:13},
which leads to an increasing number of LDA parameters that may be out of memory.
Generally,
if $W \ge 10^6$ and $K \ge 10^5$,
we require at least $400$ GBytes space to store the global topic-word parameter matrix $\boldsymbol{\hat{\phi}}_{K \times W}$.
Similarly,
the residual matrix $r_{K \times W}$ can be also processed as a parameter stream (line $15$).
In this paper,
we choose the hierarchical data format (HDF5)\footnote{\url{http://www.hdfgroup.org/HDF5/}},
which is designed for flexible and efficient I/O and for high volume and complex data.
Fault tolerance is also assured because the global topic-word matrix is stored in hard disk for restarting the online learning.

\section{Experiments} \label{s4}

\begin{table}[t]
\centering
\caption{Statistics of four document data sets.}
\begin{tabular}{|l|l|l|l|l|l|} \hline
Data sets &$D$         &$W$         &$NNZ$        &Training  &Test      \\ \hline \hline
ENRON     &$39861$     &$28102$     &$3710420$    &$37000$   &$2861$    \\ \hline
WIKI      &$20758$     &$83470$     &$9272290$    &$19000$   &$1758$    \\ \hline
NYTIMES   &$300000$    &$102660$    &$69679427$   &$290000$  &$10000$   \\ \hline
PUBMED    &$8200000$   &$141043$    &$483450157$  &$8160000$ &$40000$   \\ \hline
\end{tabular}
\label{dataset}
\end{table}

The experiments are carried out on the four publicly available data sets~\cite{Porteous:08}:
ENRON, WIKI,
NYTIMES and PUBMED in Table~\ref{dataset},
where $D$ is the total number of documents,
$W$ the vocabulary size,
and $NNZ$ the number of non-zero elements.
We randomly reserve a small proportion of documents as test sets,
and uses the remaining documents as training sets in Table~\ref{dataset}.
Our experiments are run on a single Sun Fire X4270 M2 server without parallelization.

We compare FOME with five state-of-the-art online LDA algorithms having source codes
such as OGS~\cite{Yao:09},\footnote{\url{http://mallet.cs.umass.edu/}}
OVB~\cite{Hoffman:10},\footnote{\url{http://www.cs.princeton.edu/~blei/topicmodeling.html}}
RVB~\cite{Wahabzada:11},
SOI~\cite{Mimno:12},\footnote{\url{http://mallet.cs.umass.edu/}}
and SCVB~\cite{Foulds:13}.
Note that OVB, RVB, SOI and SCVB are not designed for infinite document streams with infinite vocabulary words
because they need to know the total number of documents in the stream for scaling purposes.
In practice,
we may predefine a fixed large number for the unknown number of documents in the stream.
In SOI and RVB,
we scan or sample the same number of minibatches (documents) for training as shown in Table~\ref{dataset},
so that we can fairly compare their training convergence time with other algorithms.
For each minibatch,
when the difference of training perplexity at two successive iterations is less than $10$,
we terminate computing this minibatch and move to the next minibatch.\footnote{We practically calculate the training perplexity every $10$ iterations.}
For a fair comparison,
we transform all algorithms to Matlab MEX C platform publicly available at~\cite{Zeng:12},
and start from the same random initializations.
We repeat $5$ runs and show average results and error bars (one standard deviation)
in Figs.~\ref{foemstime},~\ref{foemsprediction},~\ref{foemtime}, and~\ref{foemprediction}.
We use their default parameters $\tau_0 = 1,024$ and $\kappa = 0.5$ in OVB, RVB, SOI
and SCVB as recommended by~\cite{Hoffman:10,Wahabzada:11,Mimno:12,Foulds:13}.
We use the predictive perplexity~\eqref{pp} on test sets as performance measures.
For all algorithms,
we use their default fixed hyperparameters $\alpha = \beta = 0.01$ in our experiments as recommended
by~\cite{Yao:09,Hoffman:10,Wahabzada:11,Mimno:12}.\footnote{VB-based algorithms need $\alpha=\beta=0.5$ as recommended in~\cite{Asuncion:09}.}
In the EM framework,
the hyperparameters $\alpha - 1 = \beta - 1 = 0.01$.

There are three main reasons for the difference of our perplexity results with those reported in
previous works~\cite{Hoffman:10,Wahabzada:11}:
1) Our vocabulary size in Table~\ref{dataset} is much larger (e.g., $W > 28,000$),
while previous works often remove less frequent words yielding a smaller vocabulary size $W < 10,000$.
Generally,
perplexity will increase with the number of vocabulary size.
2) Our held-out test sets are much larger than previous works,
which often have a higher perplexity for prediction.
3) Previous works do not show clearly the partition of the test set into two parts,
but our work shows the $80\%$ and $20\%$ partition strategy in~\eqref{pp} leading to different results.

\subsection{Dynamic Scheduling}

\begin{figure}[t]
\centering
\includegraphics[width=0.65\linewidth]{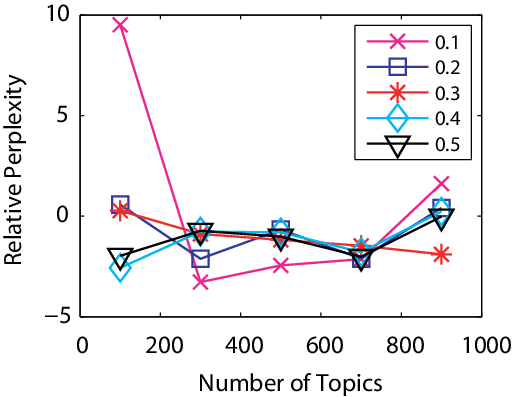}
\caption{The effectiveness of dynamic scheduling.}
\label{lambdak}
\end{figure}

We examine the parameter $\lambda_k$ of time-efficient IEM in FOEM on the relatively smaller NIPS data set,
which contains $1,500$ documents with $12,419$ vocabulary words~\cite{Porteous:08}.
We choose the training perplexity $\mathcal{P}_{\lambda_k=1}$ as the benchmark.
We change $\lambda_k$ from $0.1$ to $0.5$,
and the relative training perplexity is the difference between the training
perplexity $\mathcal{P}_{\lambda_k=\{0.1,0.2,0.3,0.4,0.5\}}$ and the benchmark.
Fig.~\ref{lambdak} shows the relative training perplexity as a function of $K$ when $\lambda_k \in \{0.1,0.2,0.3,0.4,0.5\}$.
Surprisingly,
there is no obvious difference when $\lambda_k = 0.1$ and $\lambda_k = 0.5$ especially when $K \ge 300$.
This phenomenon implies that only a small proportion of topics plays a major role when $K$ is very large.
When $\lambda_k \le 0.5$,
the time-efficient IEM achieves even a lower perplexity value than that with $\lambda_k=1$.
The reason is that most documents have very sparse responsibilities when $K$ is very large,
and thus searching the subset of topic space is enough to yield a comparable topic modeling accuracy.
Such a property as sparseness of responsibilities has been also used to speed up topic modeling~\cite{Porteous:08,Yao:09,Li:14}.
We wonder whether $\lambda_k$ can be even smaller when $K$ is very large, e.g.,
$K \in \{1500,2000\}$.
On the NIPS data set,
the time-efficient IEM with $\lambda_k = 0.05$ achieves $555.89$ and $542.70$ training perplexity, respectively.
In contrast,
the time-efficient IEM with $\lambda_k=1$ achieves $543.90$ and $533.97$ training perplexity, respectively.
The relative training perplexity is less than $2\%$.
Therefore,
it is reasonable to expect that when $K$ is very large,
$\lambda_k K$ may be a constant, e.g., $\lambda_k K = 10$.
In this case,
the training time of the time-efficient IEM will be insensitive to $K$.
This bound $\lambda_k K = 10$ is reasonable because
usually a common word is unlikely to be associated with more than $10$ topics in practice at each iteration.

\subsection{Parameter Streaming} \label{s4.2}

\begin{table}
\centering
\caption{Training time (seconds) per iteration as a function of buffer size when $K=10^4$.}
\begin{tabular}{|c|c|c|c|c|} \hline
Buffer size &$0.0$GB     &$0.2$GB     &$0.5$GB      &$0.8$GB   \\ \hline \hline
ENRON       &$5.80$      &$5.52$      &$5.30$       &$4.86$    \\ \hline
WIKI        &$21.60$     &$20.92$     &$19.80$      &$19.12$   \\ \hline
NYTIMES     &$16.80$     &$16.60$     &$16.40$      &$16.30$   \\ \hline
PUBMED      &$7.90$      &$7.72$      &$7.50$       &$6.14$    \\ \hline \hline
Buffer size &$1.0$GB     &$1.5$GB     &$2.0$GB      &in-memory \\ \hline
ENRON       &$4.18$      &$2.60$      &$2.40$       &$2.00$    \\ \hline
WIKI        &$18.84$     &$18.10$     &$17.40$      &$9.30$    \\ \hline
NYTIMES     &$16.10$     &$15.70$     &$14.80$      &$6.20$    \\ \hline
PUBMED      &$5.60$      &$3.20$      &$3.10$       &$2.40$    \\ \hline
\end{tabular}
\label{bigmodel}
\end{table}

When the size of vocabulary and the number of topics are very large,
we often cannot fit the LDA global topic-word matrix $\boldsymbol{\hat{\phi}}_{K \times W}$ in memory.
Such a big model problem has not been considered in previous online LDA algorithms.
Here,
we consider this parameter matrix as a stream.
For each minibatch at each iteration,
we load only necessary columns of parameter matrix $\boldsymbol{\hat{\phi}}^s_{K \times W_s}$ for computation.
Since the input minibatch is re-organized into a vocabulary-major sparse matrix,
we need only to perform one I/O for each minibatch at each iteration.
We may set a buffer with size $K \times W^*$ that stores parameter stream as much as possible,
which can further reduce the total frequencies of I/O.

Table~\ref{bigmodel} shows the training time per minibatch iteration as a function of buffer size when $D_s = 1,024$ and $K=10^4$.
For PUBMED,
the global parameter matrix $\boldsymbol{\hat{\phi}}_{K \times W}$ will take around $10$GB memory.
The column ``in-memory" in Table~\ref{bigmodel} shows the training time when all LDA parameters are in memory.
When we do not use the buffer,
due to high I/O frequencies,
the training time is around $3$ times slower than that of ``in-memory".
We see that when we increase the buffer size,
the training time will steadily decrease because of low I/O costs.
For ENRON and PUBMED,
each minibatch of documents contains a relatively smaller number of vocabulary words.
When the buffer size is $0.8$GB,
almost half vocabulary words in each minibatch is in buffer.
When the buffer size is $2$GB,
almost all vocabulary words in each minibatch is loaded in buffer.
As a result,
the training time at buffer size $2$GB is close to that of ``in-memory".
For WIKI and NYTIMES,
each minibatch contains a relatively more vocabulary words,
so that $2$GB buffer can hold only less than half of the vocabulary words.
This is the reason why the training time at buffer size $2$GB is still twice slower than that of in-memory.

As a summary,
our I/O strategy for LDA parameter stream is effective for big topic modeling tasks,
which is promising to handle infinite vocabulary words~\cite{Zhai:13} and the large number of topics (e.g., $K \ge 10^5$).
In our experiment,
FOME can extract $K=10^4$ topics from PUBMED using $2$GB buffer
on a common desktop computer with $4$GB memory by around one day ($29$ hours),
which cannot be done by other state-of-the-art online LDA algorithms due to high memory consumptions
for the global matrix $\hat{\boldsymbol{\phi}}_{K \times W}$.
Note that the parallel Gibb sampling algorithm on $1,024$ processors requires
approximately $50$ hours to extract $K=10^4$ topics from PUBMED~\cite{Newman:09}.
Our FOME on a single processor is even faster than this state-of-the-art large-scale parallel solution.
Given $1$TB hard disk space,
FOME is possible to extract one million topics from billions of documents based on a common desktop computer.

\subsection{Comparisons}

\begin{figure*}[h]
\centering
\includegraphics[width=0.8\linewidth]{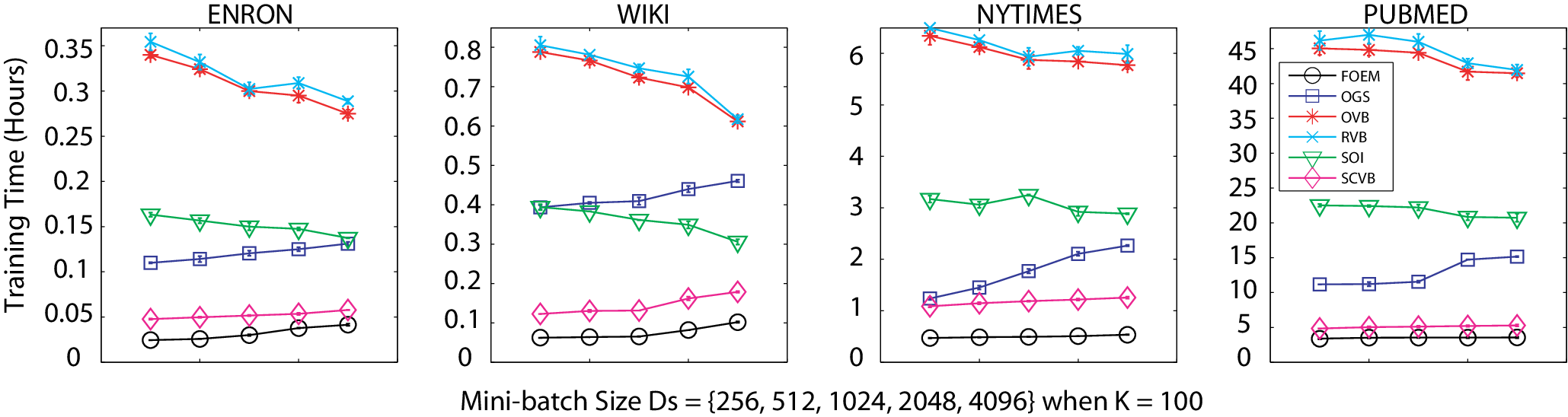}
\caption{Training convergence time (hours) as a function of the minibatch size $D_s$ when $K=100$.}
\label{foemstime}
\end{figure*}

\begin{figure*}[t]
\centering
\includegraphics[width=0.8\linewidth]{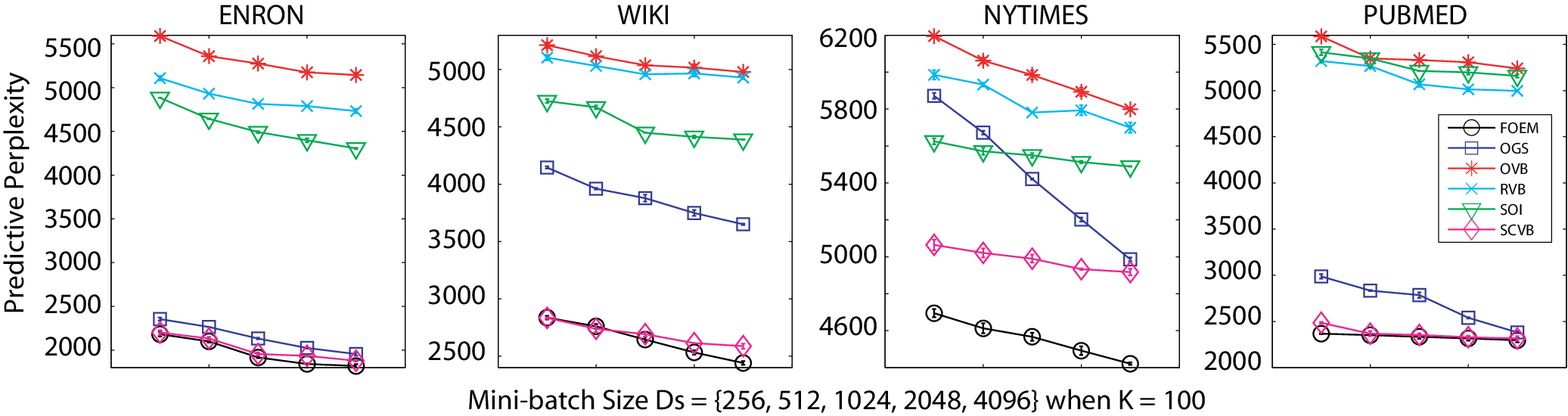}
\caption{Predictive perplexity as a function of the minibatch size $D_s$ when $K=100$.}
\label{foemsprediction}
\end{figure*}

\begin{figure*}[t]
\centering
\includegraphics[width=0.8\linewidth]{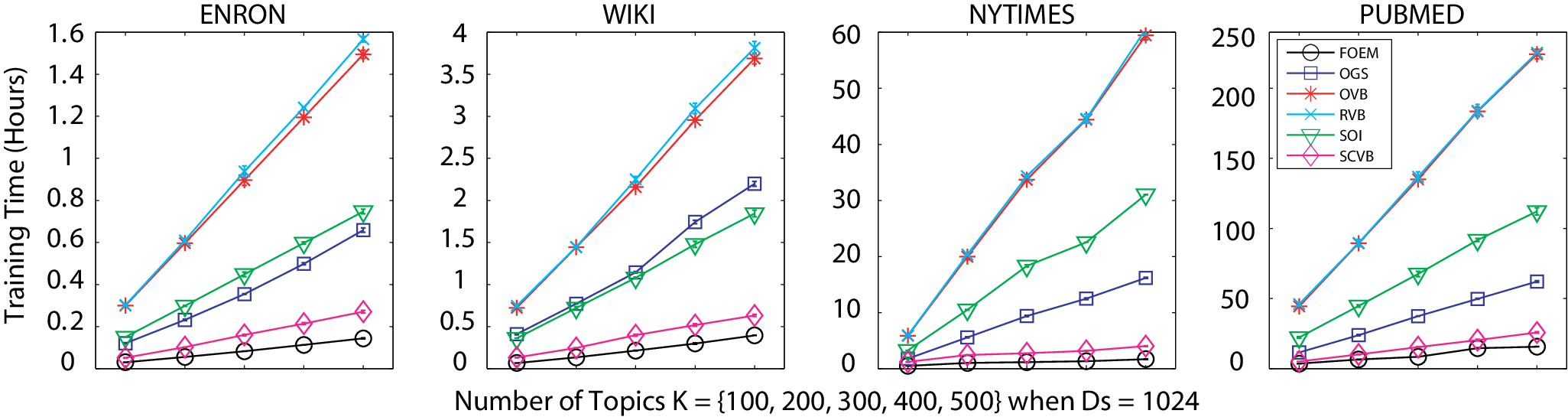}
\caption{Training convergence time as a function of the number of topics $K$ when $D_s = 1,024$.}
\label{foemtime}
\end{figure*}

\begin{figure*}[t]
\centering
\includegraphics[width=0.8\linewidth]{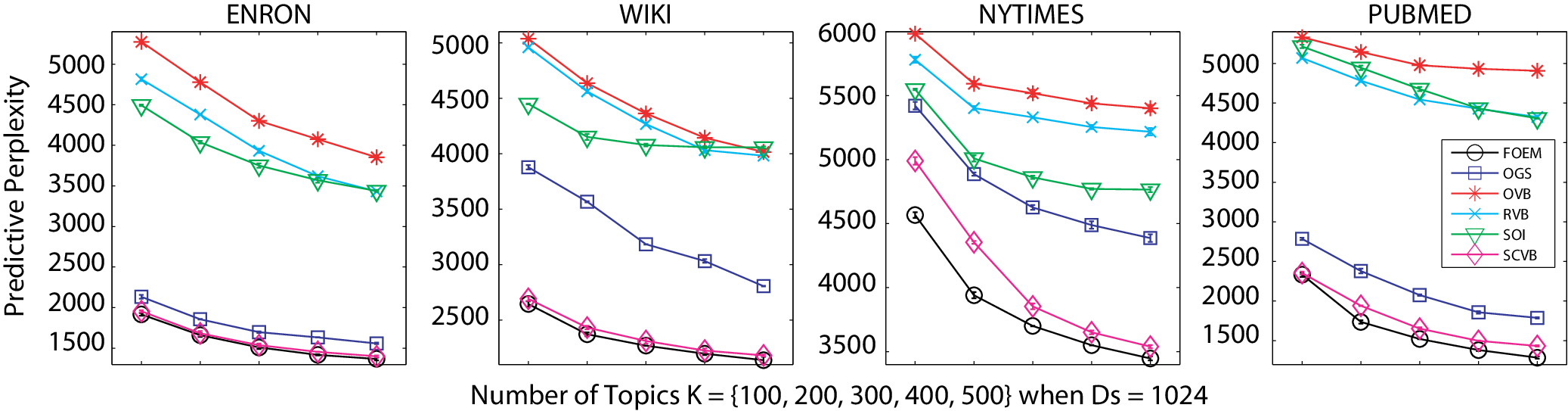}
\caption{Predictive perplexity as a function of the number of topics $K$ when $D_s = 1,024$.}
\label{foemprediction}
\end{figure*}

\begin{figure*}[t]
\centering
\includegraphics[width=0.8\linewidth]{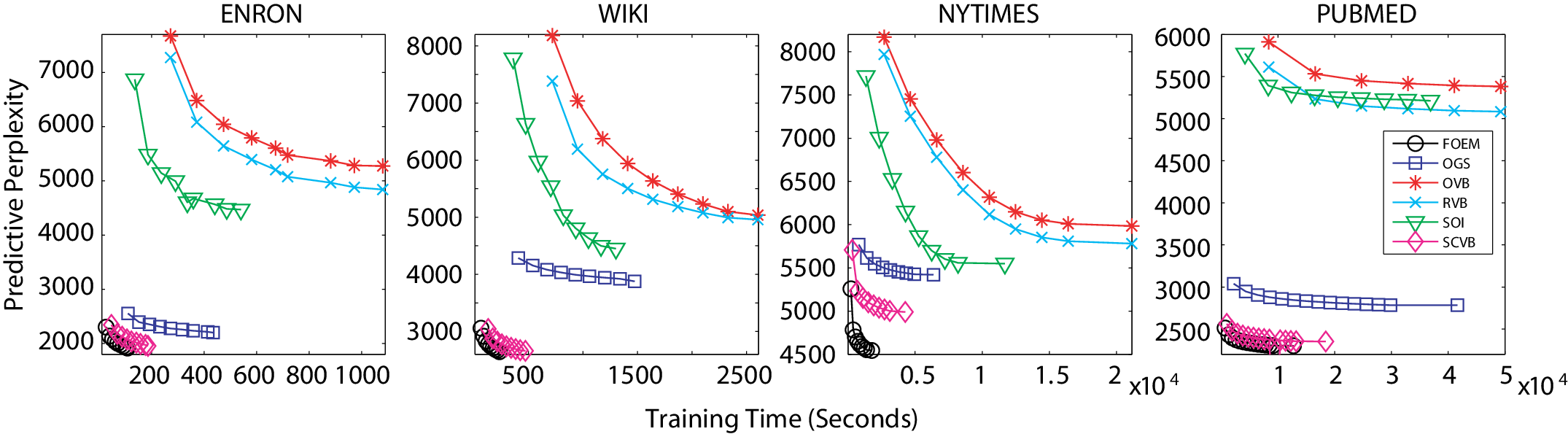}
\caption{Predictive perplexity on test set as a function of training time (seconds) when $K=100$ and $D_s=1,024$.}
\label{foemconvergence}
\end{figure*}

We compare FOEM with other state-of-the-art online LDA algorithms
in terms of the minibatch size, number of topics and convergence speed on the test set.
Fig.~\ref{foemstime} shows the training convergence time as a function of minibatch size
$D_s \in \{256,512,1024,2048,4096\}$ when $K=100$.
The training convergence time of FOME/OGS/SCVB increases slightly with the increase of the minibatch size,
while that of OVB/RVB/SOI decreases with the increase of the minibatch size.
The reason is that FOEM/OGS/SCVB (OVB/RVB/SOI) require more (less)
number of iterations for convergence when the minibatch size increases.
When $D_s \rightarrow D$,
these algorithms reduce to batch ones leading to longer (shorter) convergence time.
For the same number of minibatches,
RVB runs slightly slower than OVB because of additional dynamic scheduling cost.
When the data stream is very large,
the overall scheduling cost becomes high.
SOI uses around half of the OVB's training convergence time consistent with the results in~\cite{Mimno:12}.
OVB/RVB/SOI are slower than FOEM/OGS/SCVB partly
because they involve time-consuming digamma computations~\cite{Asuncion:09,Zeng:11}.
We see that FOME uses the least training convergence time among all algorithms.
Also,
the training time of FOME is insensitive to the minibatch size.
Fig.~\ref{foemsprediction} shows the predictive perplexity as a function of minibatch size.
All algorithms reduce the predictive perplexity when the minibatch size increases,
because the larger minibatch size will lead to more robust online gradient descents for higher topic modeling accuracy.
FOEM/OGS/SCVB have much lower predictive perplexity because they have different posterior inference
from OVB/RVB/SOI as shown in Table~\ref{complexity} and Subsection~\ref{s2.4}.
In all cases,
FOME achieves the lowest predictive perplexity showing the highest topic modeling accuracy.
However,
the larger minibatch size will consume more memory space,
so that we choose $D_s = 1,024$
to balance the memory usage and the topic modeling accuracy.

Fig.~\ref{foemtime} shows the training convergence time as a function of the number of topics
$K \in \{100, 200, 300, 400, 500\}$ when $D_s = 1,024$.
Except for FOEM,
the training convergence time of all online LDA algorithms increases linearly with the number of topics.
For the same number of minibatches,
RVB runs the slowest due to the additional scheduling cost,
which is consistent with Fig.~\ref{foemstime}.
FOME is the fastest algorithm because it is derived from the fast convergent time-efficient IEM,
and its time complexity is relatively insensitive to the number of topics $K$ in Table~\ref{complexity}.
When $K=100$,
FOEM consumes around $3.5$ hours for convergence on PUBMED.
When $K=10^4$,
FOEM consumes only $29$ hours for convergence as shown in Subsection~\ref{s4.2}.
This result confirms that FOEM's training convergence time does not increase linearly with the number of topics.
We see that OVB/RVB/SOI are slower because of time-consuming digamma functions as discussed in~\cite{Asuncion:09,Zeng:11}.
Fig.~\ref{foemprediction} shows the predictive perplexity as a function of the number of topics $K$.
We see that FOME has the lowest predictive perplexity.
Similar to Fig.~\ref{foemsprediction},
OVB/RVB/SOI have relatively higher perplexity than FOEM/OGS/SCVB because their different posterior inference
as shown in Table~\ref{complexity} and Subsection~\ref{s2.4}.

Fig.~\ref{foemconvergence} shows the predictive perplexity on test set as a function of training time.
All algorithms can converge to a stationary point by scanning more minibatches of documents.
We see two groups of algorithms having quite different convergence performances: FOME/OGS/SCVB and OVB/RVB/SOI.
The former converges faster to the lower predictive perplexity,
while the latter converges slower to the higher predictive perplexity.
The reason lies in the different posterior inference objectives in Subsection~\ref{s2.4}.
FOEM/OGS/SCVB infers the best parameter set $\{\boldsymbol{\theta},\boldsymbol{\phi}\}$
or the best labeling configuration $\mathbf{z}$,
which makes the predictive perplexity~\eqref{pp} lower
because perplexity is a function of the parameter set $\{\boldsymbol{\theta},\boldsymbol{\phi}\}$.
On the contrary,
OVB/RVB/SOI infers the best $\{\boldsymbol{\theta}, \mathbf{z}\}$ by approximation,
leading to higher predictive perplexity values.
On the four data sets,
FOEM converges around $2 \sim 5$ times faster when compared with SCVB (i.e., SEM in Fig.~\ref{sem}),
which confirms the effectiveness of the dynamic scheduling strategy used in Fig.~\ref{foem}.

\section{Conclusions} \label{s5}

This paper presents time and memory-efficient FOME for both big data streams and big LDA models on just a single PC.
We show that FOME can converge fast to the stationary point of LDA's likelihood function within the EM framework.
Extensive experiments confirm that
FOME is superior to the state-of-the-art online LDA algorithms in terms of speed, space and accuracy.
Unlike previous online algorithms,
FOEM is designed to process infinite documents with infinite vocabulary words for some lifelong topic modeling tasks.
In our future work,
we may extend and deploy FOME on the parallel multi-core
and multi-processor architectures~\cite{Newman:09,Smola:12,Liu:15} for industrial big topic modeling tasks.

\section*{Acknowledgements} \label{s6}

This work is supported by NSFC (Grant No. 61373092 and 61033013),
Natural Science Foundation of the Jiangsu Higher Education Institutions of China (Grant No. 12KJA520004),
Innovative Research Team in Soochow University (Grant No. SDT2012B02) to JZ,
and a GRF grant from RGC UGC Hong Kong (GRF Project No.9041574),
a grant from City University of Hong Kong (Project No. 7008026) to ZQL.

\end{document}